\documentclass{article}
\usepackage{ijcai21}

\usepackage{times}
\usepackage{soul}
\usepackage{url}
\usepackage[colorlinks=true,citecolor=blue]{hyperref}
\usepackage[utf8]{inputenc}
\usepackage[small]{caption}
\usepackage{graphicx}
\usepackage{amsmath,amssymb}
\usepackage{amsthm}
\usepackage{bm}
\usepackage{booktabs}
\usepackage{multirow}
\usepackage{latexsym}

\usepackage{enumerate}
\usepackage{subfigure}
\usepackage{csquotes}

\newcommand{\angstrom}{\mbox{\normalfont\AA}}

\title{A Review of Some Techniques for Inclusion of Domain-Knowledge into Deep Neural Networks}

\author{
	Tirtharaj Dash$^{*,1,3}$, Sharad Chitlangia$^{2,3}$, Aditya Ahuja$^{1,3}$, Ashwin Srinivasan$^{1,3}$
	\affiliations
	$^1$ Department of Computer Science \& Information Systems\\
	$^2$ Department of Electrical and Electronics Engineering \\
	$^3$ Anuradha and Prashanth Palakurthi Centre for AI Research (APPCAIR) \\
	BITS Pilani, K.K. Birla Goa Campus, Goa 403726, India
	\emails
	\{tirtharaj,f20170472,f20170080,ashwin\}@goa.bits-pilani.ac.in\\
	($^*$: corresponding author)
}

\begin{document}
	
\maketitle

\begin{abstract}
	We present a survey of ways in which 
	existing scientific knowledge are included when
	constructing models with neural networks.
	The inclusion of domain-knowledge is of
	special interest not just to
	constructing scientific assistants, but
	also, many other areas that involve understanding data
	using human-machine collaboration. In many
	such instances, machine-based model construction
	may benefit significantly from being provided with
	human-knowledge of the domain encoded in a sufficiently precise
	form. This paper examines the inclusion of domain-knowledge
	by means of changes to:
	the input, the loss-function,
	and the architecture of deep networks. The categorisation
	is for ease of exposition: in practice we expect a
	combination of such changes will be employed. In
	each category, we describe techniques that have
	been shown to yield significant changes in the performance of deep neural networks.
\end{abstract}

\section{Introduction}
\label{sec:intro}
\begin{figure*}[!htb]
	\centering
	\includegraphics[width=0.65\textwidth]{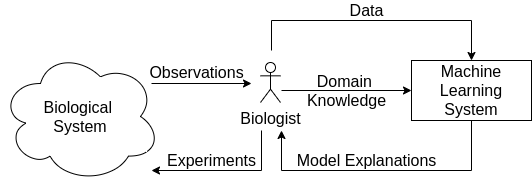}
	\caption{An example of AI for Science. The human-in-the-loop is a biologist. The biologist conducts
	experiments in a biological system, obtains
	experimental observations. The biologist then
	extracts data that can be used to construct
	machine learning model(s). Additionally,
	the machine learning system has access to
	domain knowledge that can be obtained from the
	biologist. The machine learning system then
	conveys its explanations to the biologist.
	}
	\label{fig:ml_a}
\end{figure*}

\begin{figure*}[!htb]
	\centering
	\subfigure[]{\includegraphics[width=0.46\textwidth]{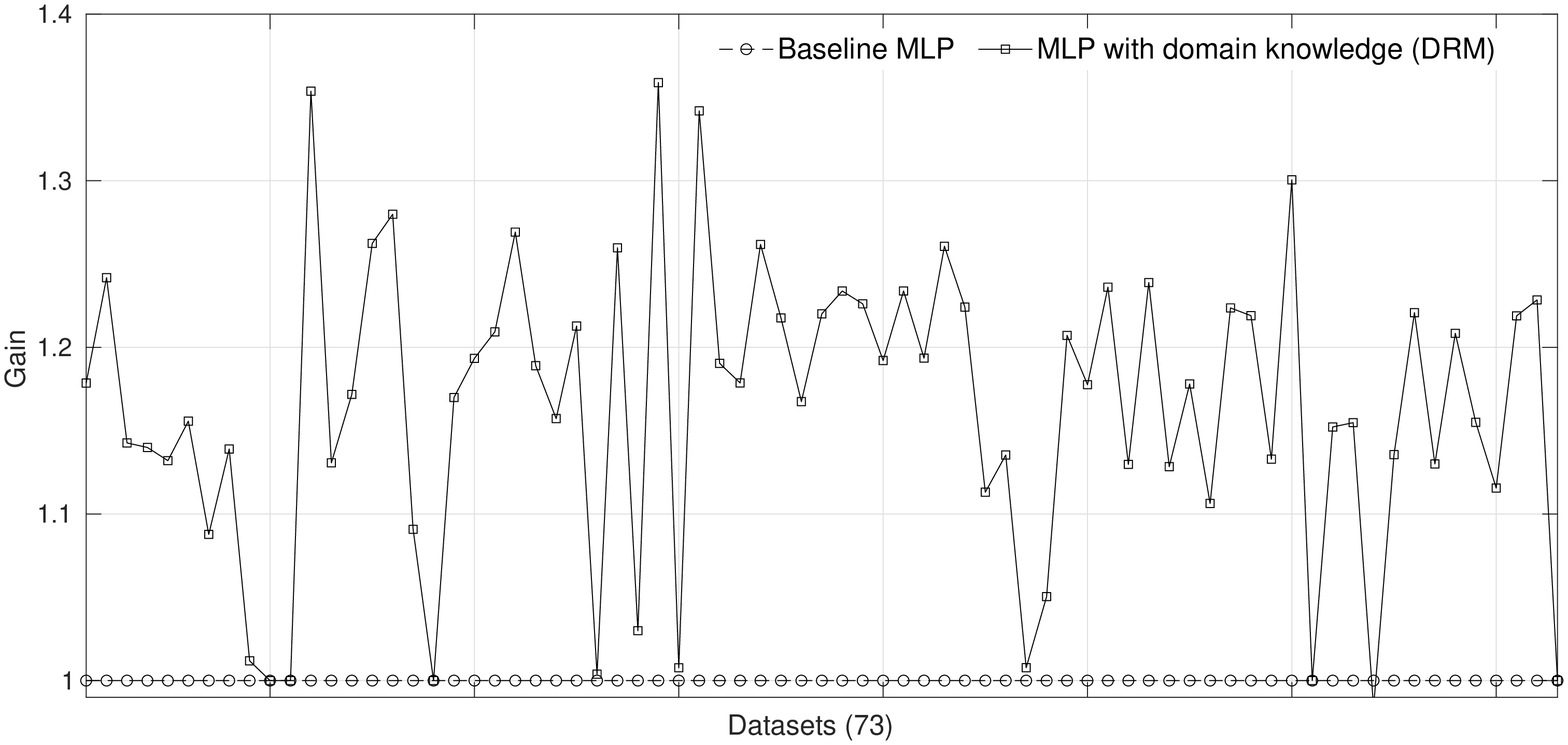}\label{fig:ml_b_mlp}} \hspace{1cm}
	\subfigure[]{\includegraphics[width=0.46\textwidth]{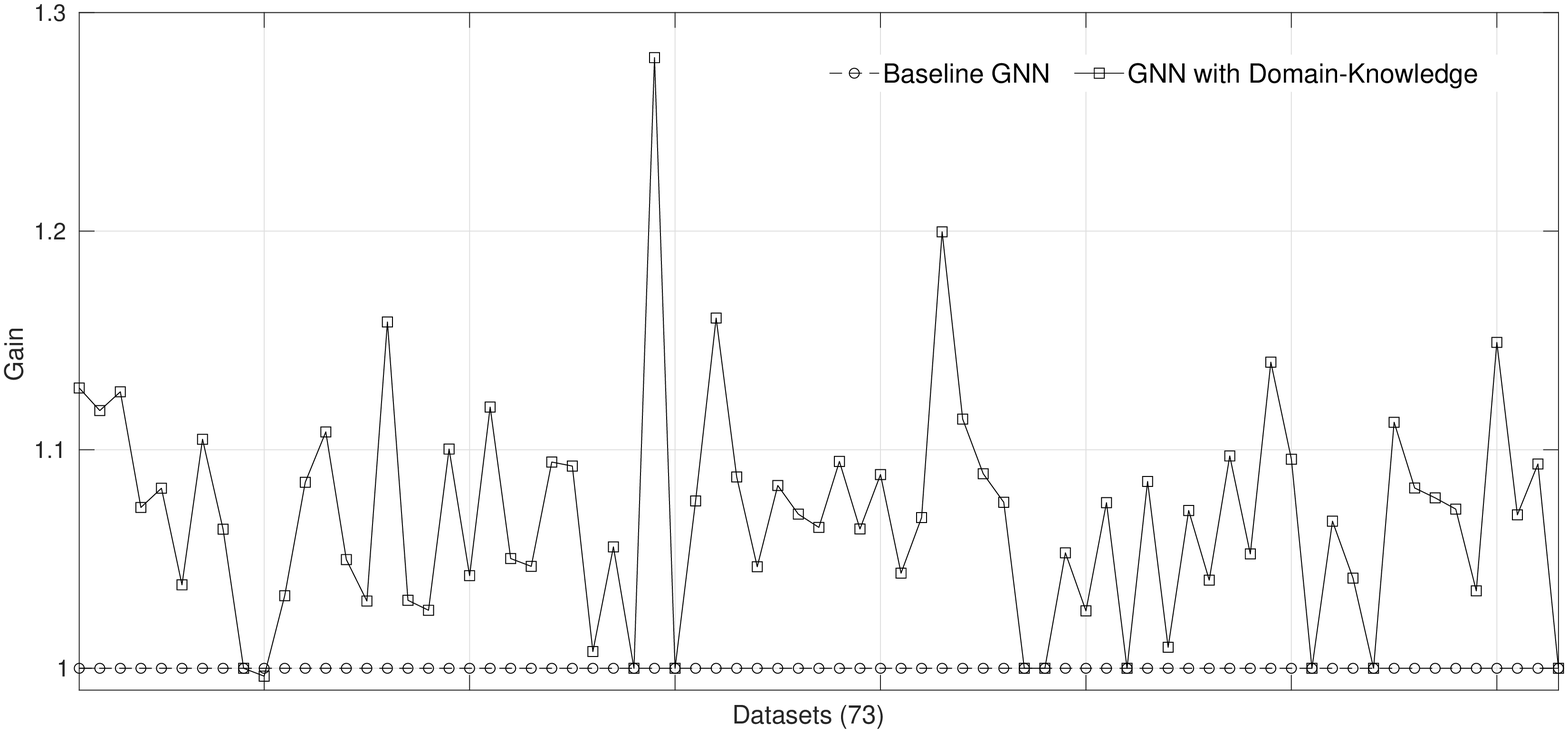}\label{fig:ml_b_botgnn}}%
	\caption{The plots   from~\protect\cite{dash2021inclusion}
		showing gains in predictive accuracy of 
		(a) multilayer perceptron (MLP),
		and (b) graph neural network (GNN) 
		with the inclusion
		of domain-knowledge. The domain knowledge
		inclusion method in (a) is a simple technique
		known as `propositionalisation'~\protect\cite{lavravc1991learning};
		and, the method in (b) is a general technique of
		incorporating domain-knowledge using bottom-graph construction.
		The results shown are over 70 datasets. 
		No importance to be given to the line joining two 
		points; this is done for visualisation purpose only.}
	\label{fig:ml_b}
\end{figure*}

Science is a cumulative enterprise,
with generations of scientists discovering, refining, correcting
and ultimately increasing our  knowledge of how things are. The accelerating pace of development in software and hardware for
machine learning--in particular, the
area of deep neural networks--inevitably
raises the prospect of
Artificial Intelligence for Science~\cite{stevens2020ai}. That is,
how can we best use AI methods
to accelerate our understanding of the natural world?
While ambitious plans
exist for completely automated AI-based robot scientists~\cite{kitano2016artificial},
the immediately useful prospect of using
AI for Science remains semi-automated. An example of such a collaborative
system is in Fig.~\ref{fig:ml_a}. For such systems
to work effectively, we need at least the following:
(1) We have to be able to tell the machine what we know,
in a suitably precise form; and (2) The machine has to
be able to tell us what it is has found, in a suitably
understandable form. While the remarkable recent successes
of deep neural networks on a wide variety of tasks makes
a substantial case for their use in model construction, it
is not immediately obvious how either (1) or (2) should be
done with deep neural networks. In this paper, we examine
ways of achieving (1). Understanding models constructed
by deep neural networks is an area of intense research
activity, and good summaries exist elsewhere \cite{lipton2016he,arrieta2019xplainable}.
To motivate the utility of providing domain-knowledge to
a deep network, we reproduce two results from \cite{dash2021incorporating}
in Fig.~\ref{fig:ml_b}, which shows that
predictive performance can
increase significantly, even with a simplified encoding
of domain-knowledge (see Fig.~\ref{fig:ml_b_mlp}).

It is unsurprising that a recent report on
AI for Science~\cite{stevens2020ai} identifies
the incorporation of domain-knowledge as one of the 3
Grand Challenges in developing AI systems:

\begin{displayquote}
	``ML and AI are generally domain-agnostic\ldots
	Off-the-shelf [ML and AI]  practice
	treats [each of these]
	datasets in the same way and ignores domain knowledge
	that extends far beyond the raw data\ldots
	Improving our ability to systematically incorporate
	diverse forms of domain knowledge can
	impact every aspect of AI.''
\end{displayquote}

But it is not just the construction of
scientific-assistants that can benefit from
this form of man-machine collaboration, and
``human-in-the-loop'' AI systems are likely
to play an increasingly important role
in engineering, medicine, healthcare,
agriculture, environment and so on \cite{tomavsev2020ai}.
In this survey, we restrict the studies on incorporation
of domain-knowledge into neural networks, with 1 or 
more hidden layers. If the domain-knowledge expressed in a symbolic form
(for example, logical relations
that are known to hold in the domain), then the 
broad area of hybrid neural-symbolic systems
(see
for example, ~\cite{garcez2012neural,raedt2020from}) is
clearly relevant to the material in this paper. However, the motivation
driving the development of hybrid systems is much broader than this paper, being
concerned with general-purpose neural-based architectures for logical representation
and inference. Here our goals are more modest: we are looking at the inclusion
of problem-specific information into machine-learning models of a kind that
will be described shortly. We 
refer the reader to \cite{besold2017neural}
for reviews of work in the broader area of neural-symbolic modelling.
More directly related to this paper is the work
on ``informed machine learning'', reviewed in ~\cite{von2021informed}.
We share with this work the interest in prior knowledge as an
important source of information that can augment existing data.
However, the goals of that paper are more ambitious than here. It aims
to identify categories of prior knowledge, using as dimensions:
the source of the knowledge, its representation, and its point of use in
a machine-learning algorithm. In this survey, we are only concerned
with some of these categories. Specifically, in
terms of the categories in \cite{von2021informed}, we are interested
in implicit or explicit sources of domain-knowledge, represented either
as logical or numeric constraints, and used at the model-construction
stage by DNNs. Informal examples of what we mean by logical and numerical
constraints are shown in Fig.~\ref{fig:constraint}.  
In general, we will assume logical constraints can
, in principle, be represented as statements in propositional logic or
predicate logic. Numerical constraints will be representable,
in principle, 
as terms in an objective function being minimised (or maximised),
or prior distributions on models. We believe this covers a
wide range of potential applications, including those
concerned with scientific discovery. 

\begin{figure}[!htb]
    \centering
    \begin{tabular}{|l|l|}
    \hline
    For inhibiting this protein:     & The model should follow that: \\
    The presence of a peroxide       & $p(y=1|\mathbf{x}) \geq 0.9$  \\
    bridge is relevant.              & $p(y=0|\mathbf{x}) \leq 0.1$  \\
    The target site is               & Initial weights should be     \\
    at most $20\angstrom$.                 & $3n - 2.3$ \cite{towell1990refinement} \\
    \multicolumn{1}{|c|}{\textbf{(a)}} & \multicolumn{1}{c|}{\textbf{(b)}} \\
    \hline
    \end{tabular}
    \caption{Informal descriptions of (a) logical;
        and (b) numerical constraints.}
    \label{fig:constraint}
\end{figure}

\subsubsection*{Focus of the Paper}

\begin{figure}[!htb]
    \centering
    \includegraphics[width=0.55\linewidth]{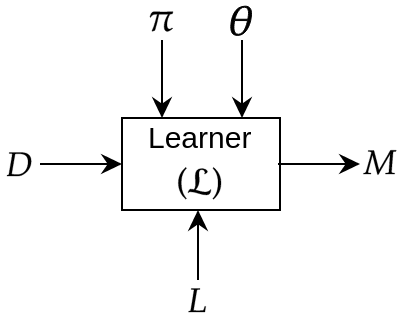}
    \caption{Construction of a deep model $M$
    from data ($D$) using a learner ($\mathcal{L}$).
    We use $\pi$ to denote the structure 
    (organisation of various layers, their interconnections etc.) 
    and $\bm{\theta}$ to denote the parameters 
    (synaptic weights) of the deep network.
    $L$ denotes the loss function (for example,
    cross-entropy loss in case of classification).
    }
    \label{fig:deepmodel}
\end{figure}

We adhere to the following informal specification
for constructing a deep neural network:
given some data $D$,
a structure and parameters of a deep network 
(denoted by $\pi$ and $\bm{\theta}$ respectively),
a learner $\mathcal{L}$ attempts to construct a neural network
model $M$ that minimises some loss function $L$.
Fig.~\ref{fig:deepmodel}
shows a diagrammatic representation.
Note that: 
(a) we do not describe how the learner $\mathcal{L}$
constructs a model $M$ given the inputs. But, it
would be normal for the  learner to optimise 
the loss $L$ by performing an iterative estimation of
the parameters $\theta$, given the structure $\pi$; and
(b) we are not concerned with how the
constructed deep model $M$ will be
used. However, it suffices to say that when
used, the model $M$ would be given one or more
data-instances encoded in the same way as was provided
for model-construction. 

In the literature, domain knowledge--also called background knowledge--does
not appear to have an accepted definition, other
than that, it refers to information about the problem. This
information can be in the form of relevant features, concepts,
taxonomies, rules-of-thumb, logical constraints, probability
distributions, mathematical distributions, causal connections and
so on. In this paper, we use the term ``domain-knowledge'' to refer
to problem-specific information that can directly be translated into
alterations to the principal inputs of Fig. \ref{fig:deepmodel}.
That is, by domain-knowledge we will mean problem-specific information that can
change:
(1) The input data to a deep network;
(2) The loss-function used; 
and (3) The model (that
is, the structure or parameters) of the deep network. 
In a sense, this progression
reflects a graded increase in the complexity
of changes involved. Figure~\ref{fig:change} tabulates
the principal implications of this position for commonly-used deep learning
architectures.

\begin{figure*}[!htb]
    \centering
    \begin{tabular}{|l|l|l|l|}
    \hline
    \multicolumn{1}{|c}{\multirow{2}{*}{\textbf{Arch.}}} &
      \multicolumn{3}{|c|}{\textbf{Domain-Knowledge Effect}} \\ \cline{2-4} 
    \multicolumn{1}{|c}{} &
      \multicolumn{1}{|c|}{\textbf{Transform Input Data}} &
      \multicolumn{1}{|c|}{\textbf{Transform Loss Function}} &
      \multicolumn{1}{|c|}{\textbf{Transform Model}} \\ \hline
    MLP &
      \begin{tabular}[c]{@{}l@{}}Reformulate\\ feature-representation\end{tabular} &
      \multirow{4}{*}{\begin{tabular}[c]{@{}l@{}}\\(For all architectures) \\
      Reformulate regularisation term;\\
      Addition of syntactic or semantic\\constraints with associated penalties;\\
      Differential costs of decisions\end{tabular}} &
      Changes in layers, hidden units \\ \cline{1-2} \cline{4-4} 
    CNN &
      \begin{tabular}[c]{@{}l@{}}Reformulate\\ spatial-representation\end{tabular} &
       &
      \begin{tabular}[c]{@{}l@{}}As with MLPs, plus changes \\ to connections between units; \\ changes convolution filters\end{tabular} \\ \cline{1-2} \cline{4-4} 
    RNN &
      \begin{tabular}[c]{@{}l@{}}Reformulate\\ sequence-representation\end{tabular} &
       &
      \begin{tabular}[c]{@{}l@{}}As with MLPs, plus possible changes \\ to attention-mechanism\end{tabular} \\ \cline{1-2} \cline{4-4} 
    GNN &
      \begin{tabular}[c]{@{}l@{}}Reformulate\\ graph-representation\end{tabular} &
       &
      \begin{tabular}[c]{@{}l@{}}As with MLPs, plus changes \\ to graph-convolution\end{tabular} \\ \hline
    \end{tabular}
    \caption{Some implications of using domain-knowledge for commonly-used deep network architectures.}
    \label{fig:change}
\end{figure*}

The rest of the paper is organised as follows:
Section~\ref{sec:input} describes inclusion of domain-knowledge by augmenting or
transformation of inputs; Section
\ref{sec:loss} describes changes that have
been employed to loss-functions; and
Section \ref{sec:model} describes
biases on parameters and changes to the
structure of deep networks.
Section~\ref{sec:challenge} outlines some major challenges
related to the inclusion of
domain-knowledge in the ways we describe.
In this section, we also present perspectives on the relevance of the use of domain-knowledge to aspects of Responsible AI, including
ethics, fairness, and explainability of DNNs.

\section{Transforming the Input Data}
\label{sec:input}

\begin{figure}[!htb]
    \centering
    \includegraphics[width=0.85\linewidth]{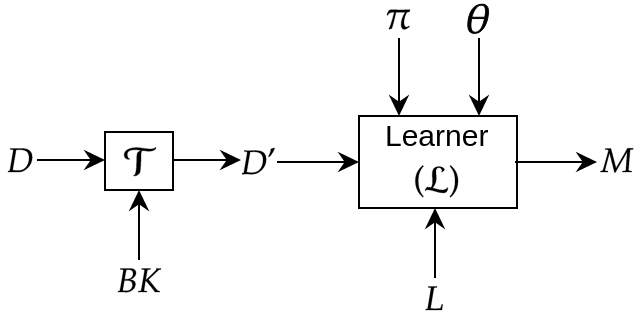}
    \caption{Introducing background knowledge
    into deep network by transforming data.
    $\mathcal{T}$ is a transformation 
    block that takes input data
    $D$, background knowledge ($BK$) and
    outputs transformed data $D'$ that is then
    used to construct a deep model using a learner
    $\mathcal{L}$.
    }
    \label{fig:bkinp}
\end{figure}

One of the prominent approaches to incorporate 
domain-knowledge into a deep network is by changing
inputs to the network. 
Here, the domain-knowledge is primarily
in symbolic form. The idea is simple: If a data
instance could be described using a set of attributes
that not only includes the raw feature-values
but also includes more details from the domain, 
then a standard deep network could then be constructed from these new features. A simple block diagram
in Fig.~\ref{fig:bkinp} shows how domain knowledge
is introduced into the network via changes in
inputs.
In this survey, we discuss broadly two different
ways of doing this: 
(a) using relational features,
mostly constructed by a method called
propositionalisation~\cite{lavravc1991learning} using another machine learning
system (for example, Inductive Logic Programming) 
that deals with data and background knowledge;
(b) without propositionalisation.

\subsection{Propositionalisation}
\label{sec:prop}

The pre-eminent form of
symbolic machine learning based on
the use of relations in first-order logic is Inductive Logic Programming (ILP)~\cite{muggleton1991inductive}, which has
an explicit role for domain-knowledge being incorporated into learning. 
The simplest use of ILP~\cite{muggleton1991inductive} to incorporate
$n$-ary relations in domain knowledge into a neural network
relies on techniques that automatically ``flatten'' 
the domain-knowledge into a set of domain-specific relational
features. 
Although not all DNNs require data to be
a set of feature-vectors, this form of data
representation is long-standing and still
sufficiently prevalent. In logical terms,
we categorise feature-based representations
as being encodings in propositional logic. The
reader would point out, correctly, that feature-values
may not be Boolean. This is correct, but we can represent
non-Boolean features by Boolean-valued propositions (for
example, a real-valued feature $f$ with value $4.2$ would
be represented by a corresponding Boolean feature $f'$ that
has the value $1$ if $f=4.2$ and $0$ otherwise). With the
caveat of this rephrasing, it has of course been possible
to provide domain-knowledge to neural networks by employing
domain-specific features devised by an expert.
However, we focus here on ways in which domain-knowledge encoded as rules in
propositional logic has been used to constrain
the structure or parameters of models constructed by a network.
Techniques for automatic construction of
Boolean-valued features from relational
domain-knowledge have a long history in the field
of ILP~\cite{mugg1994ilp,muggleton2012ilp,cropper2020turning},
originating from the LINUS~\cite{lavravc1991learning}.
Often called {\em propositionalisation\/}, the approach
involves the construction of
features that identify the conditions under which they take on
the value $1$ (or $0$). For example, given (amongst other
things) the definition of benzene rings and of fused rings,
an ILP-based propositionalisation 
may construct the Boolean-valued feature that has the value $1$ if
a molecule has 3 fused benzene rings, and $0$ otherwise.
The values of such Boolean-valued features allows us to
represent a data instance (like a molecule) as a Boolean-valued
feature-vector, which can then be provided to a neural network.
There is a long history of
propositionalisation: see \cite{Kramer2001} for a review of some
of early use of this technique, and \cite{lavrac2020prop,vig2017investi}
who examine the links between propositionalisation and
modern-day use of embeddings in neural networks.
More clearly, the authors examine that both 
propositionalisation and embedding approaches aim at
transforming data into tabular data format, while
they are being used in different problem settings and
contexts. One recent example of embedding is 
demonstrated in \cite{gaur2021can} where the authors
use different text-embedding approaches such as
sentence encoder~\cite{cer2018universal} and GPT2~\cite{radford2019language} to transform
textual domain-knowledge into embedding vectors.

A direct application of propositionalisation, demonstrating its
utility for deep networks has been its use in Deep Relational
Machines ~\cite{Lodhi2013}. A DRM is a deep 
fully-connected neural network
with Boolean-valued inputs obtained from propositionalisation
by an ILP engine. 
In \cite{dash2018large}, Boolean-valued features from an
ILP engine are sampled 
from a large space of possible features. The sampling technique
is refined further in~\cite{dash2019discrete}.

The idea of propositionalisation also
forms the foundation for a method known as
`Bottom Clause Propositionalisation (BCP)' to propositionalise
literals of a most-specific clause, or ``bottom-clause'' in ILP. Given
a data instance, the bottom-clause is 
the most-specific first-order clause that entails the data instance,
given some domain-knowledge. Loosely speaking, the most-specific
clause can be thought of ``enriching'' the data instance with
all domain relations that are true, given the
data instance. The construction of such most-specific clauses
and their subsequent use in ILP was introduced in \cite{muggleton1995inverse}. 
CILP++~\cite{francca2014fast} uses
bottom-clauses for data instances to construct feature-vectors for
neural networks. This is an extension to  CIL$^2$P~\cite{AvilaGarcez1999} where the neural network
has recurrent connections to enforce the background knowledge
during the training.

Propositionalisation has conceptual and practical limitations.
Conceptually, there is no variable-sharing
between two or more first-order logic features~\cite{dash2018large}. That is, all useful compositions have to be pre-specified. Practically, this makes
the space of possible features extremely large: this has meant
that the feature-selection has usually been done separately from
the construction of the neural network. 
In this context, 
another work that does not employ either
propositionalisation or network augmentation
considers a combination of symbolic knowledge
represented in first-order logic
with matrix factorization
techniques~\cite{rocktaschel-etal-2015-injecting}. This
exploits dependencies between textual patterns to generalise
to new relations. 

Recent work on neural-guided program
synthesis also explicitly includes
domain-specific relations. Here
programs attempt to construct
automatically compositions of functional 
primitives. The primitives are 
represented as fragments of functional programs that are
expected to be relevant.
An example of neural-guided program synthesis that uses
such domain-primitives is
DreamCoder~\cite{ellis2018dreamcoder,ellis2020dreamcoder}. 
DreamCoder receives as inputs, the partial
specification of a function
in the form of some inputs--output pairs,
and a set of low-level primitives represented in a declarative language.
Higher-level domain-concepts are then abduced as compositions
of these primitives via a neurally-guided search procedure
based on a version of the Bayesian ``wake-sleep'' algorithm~\cite{hinton1995wake}. The deep networks use
a (multi-hot) Boolean-vector encoding to represent functional
compositions (a binary digit is associated with each
primitive function, and takes the value 1 if and only if the primitive
is used in the composite function).

There are methods that do not use an explicit propositionalisation step,
but nevertheless amount to re-formulating the input feature-representation.
In the area of ``domain-adaptation''~\cite{ben2010theory}, ``source'' problems
act as a proxy for domain-knowledge for a related ``target'' problem.\footnote{
Superficially, this is also the setting underlying {\em transfer\/} learning.
However, the principal difference is that source and
target problems are closely related in domain-adaptation, but this need
not be the case with transfer-learning. Transfer-learning also usually involves
changes in both model-parameters and model-structure. Domain-adaptation does not
change the model-structure: we consider these points in a later section.} There
is a form of domain-adaptation in which the target's input representation is
changed based on the source model. In \cite{dong2021generative},  for
example, a feature-encoder ensures that the 
feature representation for the target domain that is the
same as the one used for the source.

\subsection{Binary and \textit{n}-ary Relations}
\label{sec:rel}

An influential form of representing relational domain-knowledge
takes the form {\em knowledge graph\/}, which
are labelled graphs, with vertices representing
entities and edges representing
binary relations between
entities. 
A knowledge graph provides a structure
representation for knowledge that is accessible
to both humans and machines~\cite{purohit2020knowledge}.
Knowledge graphs have been used successfully
in variety of problems arising in
information processing domains such as
search, recommendation, summarisation~\cite{sheth2019knowledge}.
Sometimes the formal semantics of knowledge graphs
such as domain ontologies are used as sources
for external domain-knowledge~\cite{yadav2021they}.
We refer the reader to \cite{hogan2020knowledge} to
a comprehensive survey of this form of representation
for domain-knowledge.

Incorporation of the information in a knowledge-graph
into deep neural models--termed ``knowledge-infused
learning''--is described in \cite{kursuncu2019nowledge,sheth2019shades}.
This aims to incorporate binary
relations contained
in application-independent sources (like DBPedia, Yago, WikiData)
and application-specific sources (like SNOMED-CT, DataMed).
The work examines techniques for incorporating relations
at various layers of deep-networks (the authors categorise
these as ``shallow'', ``semi-deep'' and ``deep'' infusion). In the case of shallow infusion,
both the external knowledge and the method
of knowledge infusion are shallow, utilising syntactic and lexical
knowledge in word embedding models. In semi-deep infusion, external knowledge is involved through attention mechanisms or learnable knowledge constraints acting as a sentinel
to guide model learning. Deep infusion employs a stratified
representation of knowledge representing different levels of abstractions in different layers of a deep learning model to transfer
the knowledge that aligns with the corresponding layer in the learning process.
Fusing the information in a knowledge-graph in this way
into various
level of hidden representations in a deep network could
also allow quantitative and qualitative assessment of its functioning, leading to knowledge-infused interpretability~\cite{gaur2021semantics}.

There have been some recent advances in introducing 
external domain-knowledge into deep sequence models.
For instance, in \cite{yadav2021they}, the authors 
incorporate domain-specific knowledge into
the popular deep learning framework, BERT~\cite{devlin2019bert} via
a declarative knowledge source like
drug-abuse ontology.
The model constructed here, called Gated-K-BERT,
is used for jointly
extracting entities and their relationship from
tweets by introducing the domain-knowledge into 
using an entity position-aware module
into the primary BERT architecture.
The experimental results demonstrate that the
incorporating domain-knowledge in this manner
leads to better relation extractors
as compared to the state-of-the-art.
This work could fall within the category of 
semi-deep infusion as described in \cite{kursuncu2019nowledge}.
\cite{yin2019domain}, in their study on
learning from electronic health records show that
the adjacency information in a medical knowledge graph can be used to model the attention mechanism in an LSTM-based RNN with attention. 
Whenever the RNN gets an entity (a medical event) 
as an input, the corresponding sub-graph in the 
medical knowledge graph 
(consisting of relations such as \textit{causes}
and \textit{is-caused-by}) is then used to 
compute an attention score. This method of incorporating
the medical relations into the RNN 
falls under the category of semi-deep knowledge
infusion.
While the above methods use the relational knowledge
from a knowledge-graph by altering or adding an
attention module within the deep sequence model,
a recent method called KRISP~\cite{marino2021krisp}
introduces such external knowledge at the output
(prediction) layer of BERT. This work could be considered
under the category of shallow infusion of domain-knowledge
as characterised by \cite{kursuncu2019nowledge}.

Knowledge graphs can be used directly by specialised
deep network models that can handle graph-based data as input
(graph neural networks, or GNNs).
The computational
machinery available in GNN then aggregates
and combines the information available in the 
knowledge graph (an example of this kind of
aggregation and pooling of relational information is
in \cite{schlichtkrull2018modeling}).
The final collected information
from this computation could be used for further
predictions. Some recent works are in~\cite{10.1145/3292500.3330855,10.1145/3308558.3313417},
where a GNN is used for estimation of
node importance in a knowledge-graph. 
The intuition is that the nodes
(in a problem involving recommender systems, as in \cite{10.1145/3308558.3313417}, a node represents
an entity) in the knowledge-graph can be represented with an aggregated
neighbourhood information with bias while 
adopting the central idea of aggregate-and-combine
in GNNs. 
The idea of
encoding a knowledge graph directly for a GNN
is also used in Knowledge-Based Recommender Dialog (KBRD)
framework developed for recommender systems~\cite{chen2019knowledgebased}.  
In this work,
the authors treat an external knowledge graph,
such as DBpedia~\cite{lehmann2015dbpedia}, as
a source of domain-knowledge allowing entities
to be enriched with this knowledge.
The authors found that the introduction of 
such knowledge in the form of a knowledge-graph
can strengthen the recommendation performance
significantly and can enhance the consistency
and diversity of the generated dialogues.
In KRISP~\cite{marino2021krisp}, a knowledge-graph
is treated as input for a GNN where each node 
of the graph network corresponds to one
specific domain-concept in the knowledge graph.
This idea is a consequence of how a GNN operates:
it can form more complex domain-concepts by propagating
information of the basic domain-concepts along the 
edges in the knowledge-graph. Further, the authors allow
the network parameters to be shared across the domain-concepts
with a hope to achieve better generalisation.
We note that while knowledge-graph provide an
explicit representation of domain-knowledge in the data,
some problems contain domain-knowledge implicitly through
an inherent topological structure (like a communication network).
Clearly, GNNs can accommodate such topological structure just in
the same manner as any other form of graph-based relations
(see for example: \cite{zhuang2019toward}).

Going beyond binary relations in knowledge-graphs and treating $n$-ary relations as hyperedges,
a technique called {\em vertex enrichment\/} is proposed in \cite{dash2021incorporating}.
Vertex-enrichment is a simplified approach  for the inclusion of domain-knowledge into
standard graph neural networks (GNNs). 
The approach incorporates first-order
background relations by augmenting
the features associated with the 
nodes of a graph provided
to a GNN. The results reported in
\cite{dash2021incorporating} show significant
improvements in the predictive accuracy of GNNs across a large number
datasets. The simplification
used in vertex-enrichment has
been made unnecessary in a recent
proposal for transforming the most-specific
clause constructed by ILP
systems employing
mode-directed inverse
entailment
(MDIE~\cite{muggleton1995inverse}).
The transformation converts this clause
into a graph can be directly used by any standard GNN ~\cite{dash2021inclusion}.
Specifically, the transformation
results in a
labelled bipartite graph consisting of
vertices representing predicates (including
domain predicates) and ground terms.
This approach reports better predictive
performance than those reported in \cite{dash2021incorporating}, and includes
knowledge-graphs as a special case. Most recently,
this method has been combined successfully 
with deep generative sequence models for generating
target-specific molecules, which
demonstrates yet another real-world
use-case of incorporating domain knowledge into
deep networks~\cite{Dash2021.07.09.451519}.

\section{Transforming the Loss Function}
\label{sec:loss}

\begin{figure}[!htb]
    \centering
    \includegraphics[width=0.6\linewidth]{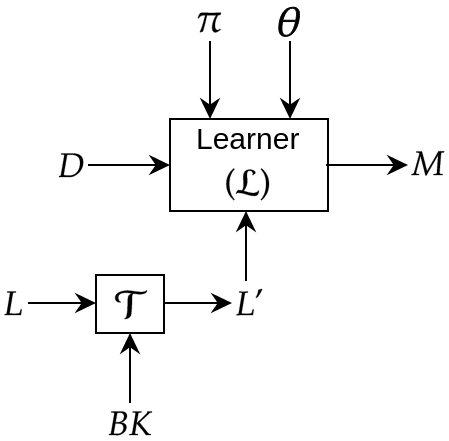}
    \caption{Introducing background knowledge 
    into deep network by 
    transforming the loss function $L$. 
    $\mathcal{T}$ block takes an input loss $L$
    and outputs a new loss function $L'$ by
    transforming (augmenting or modifying) $L$ 
    based on background knowledge ($BK$). The learner
    $\mathcal{L}$ then constructs a deep model using
    the original data $D$ and the new
    loss function $L'$.}
    \label{fig:bkloss}
\end{figure}

One standard way of incorporating domain-knowledge
into a deep network is by introducing
``penalty'' terms into the loss (or utility) function that
reflect constraints imposed by domain-knowledge.
The optimiser used for model-construction then minimises the
overall loss that includes the penalty terms.
Fig.~\ref{fig:bkloss} shows a simple block
diagram where a new loss term is introduced
based on the background knowledge.
We distinguish two kinds of domain constraints--syntactic and semantic--and
describe how these have been used to introduce penalty
terms into the loss function.

\subsection{Syntactic Loss}
\label{sec:synloss}

The usual mechanism for introducing syntactic constraints
is to introduce one or more {\em regularisation\/}
terms into the loss function. The most common form
of regularisation introduces penalties based
on model complexity (number of hidden layers, or number
of parameters and so on: see, for example,
~\cite{kukavcka2017regularization}). 

A more elaborate form of syntactic constraints involves
the concept of {\em embeddings\/}.
Embeddings refer to the relatively low-dimensional learned continuous vector
representations of discrete variables.
Penalty terms based on
``regularising embeddings'' are used to encode
syntactic constraints on the complexity of embeddings.
\cite{Fu1995} was an early work in this 
direction, in which the authors proposed a strategy to 
establish constraints by designating each node in a
Hopfield Net to represent a concept and edges to 
represent their relationships and learn these nets 
by finding the solution which maximises the greatest 
number of these constraints. \cite{rocktaschel-etal-2014-low} 
was perhaps the first method of regularising 
embeddings from declarative knowledge encoded in 
first-order logic. 
The proposal here is for mapping between
logical statements and their embeddings,
and logical inferences and matrix operations.
That is, the model behaves
as if it is following a complex first-order
reasoning process, but operates at the level
of simple vectors and matrix representations.
\cite{rocktaschel-etal-2015-injecting}
extended this to regularisation by addition of 
differentiable loss terms to the objective-based on
propositionalisation of each first-order predicate.
Guo \textit{et al.}~\cite{guo2016jointly} 
proposed a joint model, called KALE, which embeds 
facts from knowledge-graphs and logical
rules, simultaneously. Here, the facts are represented
as ground atoms with a calculated truth value
in $[0,1]$ suggesting how likely that the fact
holds. Logical rules (in grounded form) are then
interpreted as complex formulae constructed by
combining ground atoms with logical connectives,
which are then modelled by fuzzy $t$-norm operators~\cite{hajek2013metamathematics}.
The truth value that results from this operation
is nothing but a composition of the constituent
ground atoms, allowing the facts from the
knowledge graph to be incorporated into the model.

\cite{Li2020} develop a method to constraint 
individual neural layers using soft logic based 
on massively available declarative rules in ConceptNet. 
\cite{Hamilton2018EmbeddingLQ} incorporates first-order 
logic into low dimensional spaces by embedding graphs 
nodes and represents logical operators as learned 
geometric relations in the space. 
\cite{Demeester2016LiftedRI} proposed ordering of 
embedding space based on rules mined from WordNet 
and found it to better prior knowledge and 
generalisation capabilities using these relational embeddings.
\cite{10.4018/JITR.2018100109} show that domain-based
regularisation in loss function can also help in constructing deep networks with less amount of data in prediction problems concerned with cloud
computing.
In~\cite{takeishi2018knowledge},
a knowledge-based distant regularisation framework was
proposed that utilises the distance information encoded in a knowledge-graph. It defines prior distributions of model parameters using knowledge-graph embeddings. They show
that this results in an optimisation
problem for a regularised factor analysis method.

\subsection{Semantic Loss}
\label{sec:semloss}

Penalty terms can also be introduced on the
extent to which the model's prediction satisfies
semantic domain constraints. For example, 
the domain may impose specific restrictions
on the prediction (``output prediction must be in
the range $3 \ldots 6$''). One way in which such
information is provided is in the form of domain-constraints.
Penalty terms are then introduced based on the number
and importance of such constraints that are violated.

A recent work that is based on loss function is in~\cite{Xu2018}.
Here the authors propose a semantic loss that
signifies how well the outputs of the deep network
matches some given constraints encoded as
propositional rules. The general intuition
behind this idea is that the semantic loss is proportional to a negative logarithm of the probability of generating a state that satisfies the constraint when sampling values according to some probability distribution.
This kind of loss function is particularly 
useful for semi-supervised learning as these 
losses behave like self-information 
and are not constructed using explicit labels 
and can thus utilize unlabelled data. 

\cite{Hu2016HarnessingDN} proposed a framework
to incorporate first-order logic rules with the
help of an iterative distillation procedure that
transfers the structured information of logic
rules into the weights of neural networks. This is
done via a modification to the knowledge-distillation
loss proposed by Hinton et al.~\cite{hinton2015istilling}.
The authors show that taking this loss-based route
of integrating rule-based domain-knowledge allows
the flexibility of choosing a deep network architecture
suitable for the intended task.

In~\cite{Fischer2019DL2TA}, authors construct a system
for training a neural network with domain-knowledge
encoded as logical constraints. Here the available
constraints are transferred to a loss function. Specifically,
each individual logic operation (such as negation, and, or,
equality etc.) is translated to a loss term. 
The final formulation results in an optimisation problem.
The authors extract constraints on inputs that capture
certain kinds of convex sets and use them as optimisation
constraints to make the optimisation tractable. 
In the developed system, it is also possible to pose queries on the model to find inputs that satisfy a set of constraints. In a similar line, \cite{muralidhar2019incorporating} proposed 
domain-adapted neural network (DANN) that works with a balanced
loss function at the intersection of models based on purely domain-based loss or purely inductive loss. Specifically,
they introduce a domain-loss term that requires a functional
form of approximation and monotonicity constraints on the
outputs of a deep network. Without detailing
much on the underlying equations, it suffices to say that formulating
the domain loss using these constraints enforces the model
to learn not only from training data but also in accordance 
with certain accepted domain rules.

Another popular approach that treats domain
knowledge as `domain constraints' is
semantic based regularisation~\cite{diligenti2017semantic,diligenti2017integrating}. 
It builds standard multilayered neural networks
(e.g. MLP) with kernel machines at the input
layer that deal with continuous-valued features.
The output of the kernel machines is input
to the higher layers implementing a fuzzy
generalisation of the domain constraints that
are represented in first-order logic.
The regularisation term, consisting of
a sum of fuzzy generalisation of constraints
using $t$-norm operations, in the
cumulative loss then penalises each violation of the constraints during the training of the deep network.
\cite{silvestri2020injecting} inject
domain knowledge at training time
via an approach that combines semantic
based regularisation and 
constraint programming~\cite{rossi2006handbook}.
This approach uses the concept of 
`propagators', which is inherent in
constraint programming to identify
infeasible assignments of variables to values
in the domain of the variables.
The role of
semantic-based regularisation is to then 
penalise these infeasible assignments weighted
by a penalty parameter.
This is an example of constraints on inputs.
In a similar line, \cite{luo2021deep} introduce domain-knowledge
into a deep LSTM-based RNN at three different levels:
constraining the inputs by designing a filtering module based
on the domain-specific rules,
constraining the output by enforcing an output range,
and also by introducing a penalty term in the loss function.

A library for integrating symbolic domain-knowledge
with deep neural networks was introduced recently in \cite{faghihi2021domiknows}. It provides some
effective ways of specifying domain-knowledge,
albeit restricted to (binary) hierarchical concepts only,
for problems arising in
the domain of natural language
processing and some subset of computer vision.
The principle of integration involves constraint
satisfaction using a primal-dual formulation
of the optimisation problem. That is: 
the goal is to satisfy the maximum number of domain constraints
while also minimising the training loss,
an approach similar to the idea proposed in 
\cite{Fischer2019DL2TA,muralidhar2019incorporating,silvestri2020injecting}.

While adding a domain-based constraint term to the loss function may seem appealing, there are a few challenges.
One challenge that is pointed out in a recent study~\cite{hoernle2021multiplexnet} is that 
incorporating domain-knowledge in this manner
(that is: adding a domain-based loss to the 
standard problem-specific loss)
may not always be suitable while dealing with
safety-critical domains where 100\% constraint 
satisfaction is desirable. One way to guarantee 100\% domain-constraint satisfaction is by directly
augmenting the output layer with some transformations 
and then deriving a new loss function due to these
transformations. These transformations are such that
they guarantee the output of the network to
satisfy the domain constraints. 
In this study, called MultiplexNet~\cite{hoernle2021multiplexnet}, the domain-knowledge
is represented as a logical formula in disjunctive normal form (DNF)
Here the output (or prediction) layer of a deep network
is viewed as a multiplexor in a logical circuit that permits
branching in logic. That is, the output of the network
always satisfies one of the constraints
specified in the domain knowledge (disjunctive formula).

The other form of semantic loss could be one that
involves a human for post-hoc
evaluation of a deep model constructed from a
set of first-order rules.
In this line, \cite{honda2021analogical}
proposed an analogical reasoning system 
intended for discovering rules by training a sequence-to-sequence model using a training
set of rules represented in first-order logic.
Here the role of domain-knowledge comes post
training of the deep sequence model;
that is, an evaluator (a human expert) tests each discovered
rule from the model by unifying them against
the (domain) knowledge base. The domain-knowledge
here serves as some kind of a validation set
where if the ratio of
successful rule unification crosses a certain threshold, then the set of discovered rules are
accepted.

\section{Transforming the Model}
\label{sec:model}

\begin{figure*}[!htb]
	\centering
	\subfigure[]{\includegraphics[width=0.25\linewidth]{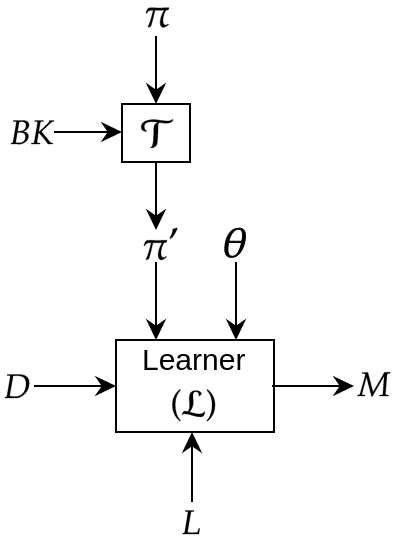}\label{fig:strbk}} \hspace{1cm}
	\subfigure[]{\includegraphics[width=0.25\linewidth]{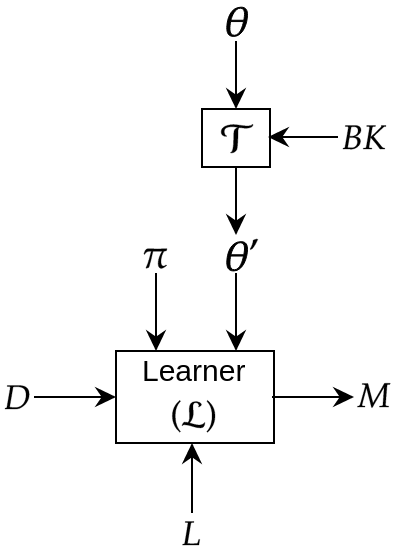}\label{fig:parambk}}%
	\caption{Introducing background knowledge
	into deep network by transforming the model
	(structure and parameter).
	In (a), the transformation block $\mathcal{T}$
	takes a input structure of a model $\pi$ and
	outputs a transformed structure $\pi'$ based 
	on background knowledge ($BK$).
	In (b), the transformation block $\mathcal{T}$
	takes a set of parameters $\bm{\theta}$ of a model 
	and
	outputs a transformed set of parameters $\pi'$ 
	based 
	on background knowledge ($BK$).
	}
	\label{fig:bkmodel}
\end{figure*}

Over the years, many studies have shown that
domain knowledge can be incorporated into a 
deep network by introducing constraints on the 
model parameters (weights) or by making a design
choice of its structure. Fig.~\ref{fig:bkmodel} 
shows a simple block diagram showing 
domain knowledge incorporation at the design stage
of the deep network.

\subsection{Constraints on Parameters}
\label{sec:param}

In a Bayesian formulation, there is an explicit mechanism for
the inclusion of domain-knowledge through the use of priors.
The regularisation terms in loss-functions, for example, can
be seen as an encoding of such prior information, usually
on the network's structure. Priors can also be introduced
on the parameters (weights) of a network. Explicitly, these would
take the form of a prior distribution over the values
of the weights in the network.
The priors on networks and network weights represent our expectations about networks before receiving any
data, and correspond to penalty terms or regularisers.
Buntine and Weigend~\cite{Buntine1991BayesianB}
extensively study how Bayesian theory can be
highly relevant to the problem of training feedforward
neural networks. This work is explicitly concerned with
choosing an appropriate
network structure and size based on prior domain-knowledge
and with selecting a prior for the weights. 

The work by \cite{hintonnealthesis} on Bayesian learning for neural networks also showed how domain-knowledge could help build a prior probability distribution over neural network parameters. In this, the shown
priors allow networks to be ``self-regularised'' to not over-fit even when the complexity of 
the neural network is increased.
In a similar spirit,
\cite{Krupka2007IncorporatingPK} showed how prior domain
knowledge could be used to define `meta-features'
that can aid in defining the prior distribution of weights.
These meta-features are additional information about
each of the features in the available data. For instance,
for an image recognition task, the meta-feature could be
the relative position of a pixel $(x, y)$ in the image. 
This meta information can be used to construct a prior
over the weights for the original features.

\subsubsection{Transfer Learning}

Transfer Learning is a mechanism to
introduce priors on weights when data
is scarce for a problem (usually
called the ``target'' domain). Transfer
learning relies on data availability for a problem similar to the target domain (usually
called the ``source'' domain). From the position taken in this paper,
domain-knowledge for transfer learning is used to change the
structure or the parameter values (or both) for a model for the target problem.
The nature of this domain-knowledge can be seen prior distributions
on the structure and/or parameter-values (weights) of models for
the target problem. The prior distributions for the target model are obtained
from the models constructed for the source problem.

In practice, transfer learning from a source domain to a 
target domain usually involves a transfer
of weights from models constructed for the source domain to the network in the target domain. This has been shown to boost performance significantly.
From the Bayesian perspective, transfer learning
allows the construction of the prior over the weights
of a neural network for the target domain based on
the posterior constructed in the source domain. 
Transfer learning is not limited by the kind of task (such as 
classification, regression, etc.) but rather by the availability of related problems. 
Language models
are some of the very successful examples of the use
of transfer learning, where the models are
initially learnt on a huge corpus of data and fine-tuned 
for other more specialised tasks.
\cite{zhuang2020comprehensive} provides
an in-depth review of some of the mechanisms and the strategies of transfer learning. Transfer learning need not be
restricted to deep networks only: in a recent study, 
\cite{liu2018mproving} proposes a model that transfers
knowledge from a neural network to a decision tree
using knowledge distillation framework. The symbolic
knowledge encoded in the decision tree could further
be utilised for a variety of tasks.

A subcategory of transfer learning is one in which the problem
(or task) remains the same, but there is a change in the distribution
over the input data from the source and 
the target. This form of learning is viewed as an instance of
domain-adaptation~\cite{ben2010theory}. Similar to
transfer learning, the knowledge is transferred
from a source domain to a target domain
in the form of a prior distribution over the model parameters.
This form of domain-adaptation uses the same model structure as the source, along
with an initial set of parameter values obtained from the source model.
The parameter values are then fine-tuned using
labelled and unlabelled data from the target data
\cite{tzeng2015simultaneous}. 
An example of this kind of learning is in \cite{du2020adversarial} where a BERT model is fine-tuned with data from multiple domains.
There are some recent surveys along these lines: \cite{wang2018deep,ramponi2020neural}.

\subsection{Specialised Structures}
\label{sec:struc}

DNN based methods arguably work best if
the domain-knowledge is used to inspire their
architecture choices~\cite{berner2021modern}.
There are reports on incorporating 
first-order logic
constructs into the structure of the
network. This allows neural-networks
to operate directly on the
logical sentences comprising
domain-knowledge.

Domain-knowledge encoded as a set
of propositional rules are used to  
constrain the structure of the neural network.
Parameter-learning (updating of the
network weights) then proceeds as normal, using
the structure. The result could be thought of
as learning weighted forms of the antecedents
present in the rules.
The most popular and oldest work along this line
is Knowledge-Based Artificial Neural Network (KBANN)~\cite{towell1990refinement}
that incorporates knowledge into neural networks. 
In KBANN, the domain knowledge is 
represented as a set of hierarchically structured propositional rules that directly determines a
fixed topological structure of a neural network~\cite{towell1994knowledge}. 
KBANN was successful in many real-world applications; but,
its representational power was bounded by
pre-existing set of rules which restricted it to
refine these existing rules rather than discovering new
rules. A similar study is KBCNN~\cite{fu1993knowledge},
which first identifies and links
domain attributes and concepts consistent with
initial domain knowledge.
Further, KBCNN introduces additional hidden
units into the network and most importantly,
it allowed decoding of the learned rules from
the network in symbolic form. However, both
KBANN and KBCNN were not appropriate for learning
new rules because of the way the initial structure
was constructed using the initial domain knowledge base.

Some of the limitations described above could be
overcome with the proposal of a hybrid system
by Fletcher and Obradovic~\cite{fletcher1993combining}.
The system was able to learn a neural
network structure that could construct new rules from
an initial set of rules. Here, the domain knowledge is
transformed into an initial network through an
extended version of KBANN's symbolic knowledge encoding.
It performed incremental hidden unit
generation thereby allowing construction or
extension of initial rule-base. In a similar manner,
there was a proposal for using Cascade ARTMAP~\cite{Tan1997} which could not only 
construct a neural network structure from 
rules but also perform explicit cascading of rules
and multistep inferencing. It was found that 
the rules extracted from Cascade ARTMAP
are more accurate and much cleaner than the rules
extracted from KBANN~\cite{towell1993extracting}.

In the late 1990s, Garcez and Zaverucha proposed
a massively parallel computational model
called CIL$^2$P based on
feedforward neural network that integrates
inductive learning from examples and domain knowledge,
expressed as a propositional logic program~\cite{AvilaGarcez1999}.
A translation algorithm generates a neural network.
Unlike KBANN, the approach uses the notion of 
``bipolar semi-linear'' neurons. This allows the
proof of a form of correctness, showing the existence of
a neural-network structure that can compute the logical
consequences of the domain-knowledge.
The output of such a network, when combined into
subsequent processing naturally incorporates the intended
interpretation of the domain predicates.
The authors extend this to the use of first-order logic
programs: we have already considered this in Sec.~\ref{sec:input}.


A recent proposal focuses on embedding symbolic
knowledge expressed as logical rules~\cite{Xie2019}.
It considers two languages of representations:
Conjunctive Normal Form (CNF) and
decision-Deterministic Decomposable Negation
Normal form (d-DNNF), which can naturally 
be represented as graph structures. The graph
structures can be provided to a graph neural network (GNN) to
learn an embedding suitable for further
task-specific implementations. 

Somewhat in a similar vein to the work
by \cite{AvilaGarcez1999}, the work reported
in \cite{Xu2018} considers
as a set of propositional statements representing
domain constraints. A deep
network is then trained to find satisfying assignments
for the constraints. Again, once such a network is constructed,
it can clearly be used in subsequent processing, capturing
the effect of the domain constraints. The network is
trained using a semantic loss that we 
have described in Sec.~\ref{sec:semloss}.

In \cite{Li2020} it is proposed to augment a language model that uses a
deep net architecture with additional statements in first-order logic.
Thus, given domain-knowledge encoded as first-order relations, connections are introduced into the network based on the logical constraints enforced by the domain-relations. The approach is related somewhat to the
work in \cite{DBLP:journals/jair/SourekAZSK18} that does not explicitly consider the incorporation
of domain-knowledge but does constrain a deep neural network's structure by first grounding a set of
weighted first-order definite
clauses and then turning them into propositional programs.

We note that newer areas are emerging that use representations
for domain-knowledge that go beyond first-order logic relations.
This includes probabilistic first-order logic, as a way of including
uncertain domain-knowledge~\cite{manhaeve2018deepproblog}. 
One interesting way this is being used is to constrain the training of  ``neural predicates'', which represent probabilistic relations that are implemented
by neural networks, and the framework can be
trained in an end-to-end fashion~\cite{manhaeve2018deepproblog,Winters2021DeepStochLogNS}. In DeepProbLog~\cite{manhaeve2018deepproblog}, for example, high-level logical reasoning can be combined
with the sub-symbolic discriminative power of deep
networks. For instance, a logic program for adding
two digits and producing the output sum is 
straightforward. However, what if the inputs 
are images of the corresponding digits? Here, 
a deep network is used to map an image to a digit,
while a (weighted) logic program, written
in ProbLog~\cite{de2007problog}, for the addition operation is treated as
the symbolic domain knowledge.
The ProbLog program is extended with a set of
ground neural predicates for which the weights
correspond to the probability distribution of
classes of digits (0 \ldots 9). The parameters
(weights of predicates and weights of neural network)
are learned in an end-to-end fashion.
A recent approach called DeepStochLog~\cite{Winters2021DeepStochLogNS} is 
a framework that extends the idea of neural 
predicates in DeepProbLog to definite clause
grammars~\cite{pereira1980definite}.
The reader may note that although DeepProbLog
and DeepStochLog do not really transform the
structure of the deep network, 
we are still considering these methods
under the heading of specialised structures.
This is because of the fact that the hybrid architecture
is a tightly coupled approach combining
probabilistic logic and deep neural networks.

One of the approaches involves
transformation of a probabilistic logic program to 
graph-structured representation. For instance, in kLog~\cite{frasconi2014klog} the
transformed representation is an undirected bipartite graph in the form of 
`Probabilistic Entity-Relationship model' \cite{heckerman2007probabilistic} which allows the use of a
graph-kernel~\cite{vishwanathan2010graph} for data classification purpose,
where each data instance is represented as a
logic program constructed from data and
background-knowledge. Another approach uses weighted
logic programs or \textit{templates} with GNNs \cite{sourek2020beyond}
demonstrating how simple relational logic programs can capture advanced graph convolution operations in a tightly integrated manner. However, it requires the use of a language of Lifted Relational Neural
Networks (LRNNs)~\cite{sourek2018lifted}. 
Template-based construction of deep network structure can 
be also seen in Logical Neural Networks (LNNs:~\cite{riegel2020logical}). 
LNNs resemble a tree structure where the leaf nodes represent
the facts in the data and background knowledge, the internal
nodes implement logical connectives (and, or, implication, etc.) using $t$-norm operators derived from real-valued logic, 
and the outputs represent the rules~\cite{sen2021neuro}. 
The inputs to LNNs are 
instantiated facts (Boolean), and the network is trained by
minimising a constrained loss function which is a consequence
of such a specialised structure.

An interesting proposal is to transform facts and rules,
all represented in (weighted) first-order logic into
matrix (or tensor) representations. 
Learning and inference can then
be conducted on these matrices (or tensors)~\cite{serafini2016logic,cohen2020tensorlog}
allowing faster computation.
NeuralLog~\cite{guimaraes2021neurallog}, for example,
extends this idea and constructs a multilayered
neural network, to some extent, 
similar to the ones in LRNN
consisting of fact layer, rule layer and 
literal layer etc. The learning here refers
to the updates of the weights of the rules. Another
work that translates domain-knowledge in 
first-order logic into a deep neural network architecture
consisting of the input layer (grounded atoms), 
propositional layer, quantifier layer and 
output layer is ~\cite{diligenti2017semantic}.
Similar to LRNN, it uses the fuzzy $t$-norm operators
for translating logical OR and AND operations.

Further emerging areas look forward to  
providing domain-knowledge as higher-order
logic templates (or ``meta-rules'': see \cite{cropper2020turning} 
for pointers to this area). 
To the best of our knowledge, 
there are, as yet, no reports in the literature on 
how such higher-order statements can
be incorporated into deep networks.

\begin{table*}[!htb]
	\centering
	\begin{tabular}{|c|c|c|}
	\hline
	\textbf{Principal Approach} &
	    \textbf{Work (Reference)} &
	    \textbf{Type of Learner} \\
	\hline
	Transforming Data &
	    DRM~\cite{Lodhi2013,dash2018large} & 
	    MLP \\
	&
	    CILP++~\cite{francca2014fast} & 
	    MLP \\
	&
	    R-GCN~\cite{schlichtkrull2018modeling} &
	    GNN \\
	&
	    KGCN~\cite{10.1145/3308558.3313417} &
	    GNN \\
	&   
	    KBRD~\cite{chen2019knowledgebased} &
	    GNN \\
	&
	    DG-RNN~\cite{yin2019domain} &
	    RNN \\
	&
	    DreamCoder~\cite{ellis2020dreamcoder} &
	    DNN$^*$ \\
	&
	    Gated-K-BERT~\cite{yadav2021they} &
	    Transformer \\
	&
	    VEGNN~\cite{dash2021incorporating} & 
	    GNN \\
	&
	    BotGNN~\cite{dash2021inclusion} & 
	    GNN \\
	&
	    KRISP~\cite{marino2021krisp} &
	    GNN, Transformer \\
	\hline
	Transforming Loss &
	    IPKFL~\cite{Krupka2007IncorporatingPK} & 
	    CNN \\
	&
	    ILBKRME~\cite{rocktaschel-etal-2015-injecting} & 
	    MLP \\
	&
	    HDNNLR~\cite{Hu2016HarnessingDN} & 
	    CNN, RNN\\
	&
	    SBR~\cite{diligenti2017semantic} & 
	    MLP \\
	&
	    SBR~\cite{diligenti2017integrating} & 
	    CNN \\
	&
	    DL2~\cite{Fischer2019DL2TA} & 
	    CNN \\
	&
	    Semantic Loss~\cite{Xu2018} & 
	    CNN\\
	&
	    LENSR~\cite{Xie2019} & 
	    GNN \\
	&
	    DANN~\cite{muralidhar2019incorporating} &
	    MLP \\
	&
	    PC-LSTM~\cite{luo2021deep} &
	    RNN \\
	&
	    DomiKnowS~\cite{faghihi2021domiknows} &
	    DNN* \\
	&
	    MultiplexNet~\cite{hoernle2021multiplexnet} &
	    MLP, CNN \\
	&
	    Analogy Model~\cite{honda2021analogical} &
	    RNN \\
	\hline
	Transforming Model &
	    KBANN~\cite{towell1994knowledge} &
	    MLP \\
	&
	    Cascade-ARTMAP~\cite{Tan1997} & 
	    ARTMAP \\
	&
	    CIL$^2$P~\cite{AvilaGarcez1999} & 
	    RNN \\
	&
	    DeepProbLog~\cite{manhaeve2018deepproblog} &
	    CNN \\
	&
	    LRNN~\cite{sourek2018lifted} &
	    MLP \\
	&
	    TensorLog~\cite{cohen2020tensorlog} &
	    MLP \\
	&
	    Domain-Aware BERT~\cite{du2020adversarial} &
	    Transformer \\
	&
	    NeuralLog~\cite{guimaraes2021neurallog} &
	    MLP \\
	&
	    DeepStochLog~\cite{Winters2021DeepStochLogNS} &
	    DNN* \\
	&
	    LNN~\cite{sen2021neuro} &
	    MLP \\
	\hline
	\end{tabular}
	\caption{Some selected works, in no particular order, showing the principal approach
	of domain knowledge inclusion into deep
	neural networks. For each work referred here,
	we show the type of learner with following acronyms:
	Multilayer Perceptron (MLP),
	Convolutional Neural Network (CNN), 
	Recurrent Neural Network (RNN), 
	Graph Neural Network (GNN),
	Adaptive Resonance Theory-based Network Map (ARTMAP),
	DNN$^*$ refers to a DNN structure dependent on intended
	task. We use `MLP' here to represent
	any neural network, that 
	conforms to a layered-structure that may or
	maynot be fully-connected. RNN also refers
	to sequence models constructed using Long Short-Term Memory (LSTM) or Gated Recurrent Unit (GRU) cells.}
	\label{tab:summary}
\end{table*}

\section{Challenges and Concluding Remarks}
\label{sec:challenge}

We summarise our discussion on domain-knowledge
as constraints in Table~\ref{tab:summary}.
We now outline some challenges in incorporating
domain-knowledge encoded as logical or numerical
constraints into a deep network. We first outline
some immediate practical challenges concerning the
logical constraints:
\begin{itemize}
	\item There is no standard framework for translating logical constraints to neural networks.
	While there are simplification methods which
	first construct a representation of the logical
	constraint that a standard
	deep network can consume, this process has its limitations
	as described in the relevant section above.
	\item Logic is not differentiable. This does not allow using standard training of deep network using
	gradient-based methods in an end-to-end fashion. 
	Propagating gradients via logic has now been looked at in \cite{evans2018delILP}, but the solution is intractable and does not allow day-to-day use.
	\item Many neural network structures are directed acyclic graphs (DAGs).
	However, transforming logical formulae directly
	into neural network structures in the manner
	described in some of the discussed works can introduce cyclic dependencies, which may need a separate form of translations.
\end{itemize}
There are also practical challenges concerning the numerical constraints:
\begin{itemize}
	\item We have seen that the numerical constraints are
	often provided with the help of modification to 
	a loss function. Given some domain-knowledge in a 
	logical representation, constructing a term in loss function is not straightforward. 
	\item Incorporating domain-knowledge via domain-based
	loss may not be suitable for some safety-critical
	applications.
	\item The process of introducing a loss term
	often results in a difficult optimisation problem
	(sometimes constrained) to be solved. This may
	require additional mathematical tools for a solution that can be implemented practically.
	\item Deep network structures constrained via 
	logical domain-knowledge may not always be scalable to large datasets.
\end{itemize}

It is possible to consider representing domain-knowledge not as logical
or numeric constraints, but through statements in natural language.
Recent rapid progress in the area of language models, for example,
the models based on attention~\cite{vaswani2017attention,brown2020language}
raises the possibility of incorporating 
domain-knowledge through conversations. While the precision of these
formal representations may continue to be needed for the construction
of scientific assistants, the flexibility of natural language may be
especially useful in communicating commonsense
knowledge to day-to-day machine assistants 
that need to an informal knowledge of the world~\cite{tandon2018commonsense,zhao2021ethical}.
Progress in this is being made (see, for example, \url{https://allenai.org/aristo}), but there
is much more that needs to be done to make the language models
required accessible to everyday machinery.

More broadly, incorporating domain-knowledge
into learning
is highlighted in \cite{stevens2020ai} as one of the Grand Challenges
facing the foundations of AI and ML. Addressing this challenge effectively
is seen as being relevant to issues arising in
automated model-construction like data-efficiency and constraint-satisfaction.
Additionally, it is suggested that developing a mapping of
internal representations of the model to domain-concepts maybe 
necessary for acceptable explanations for the model's predictions and for
developing trust in the model.

It is now accepted that trust comes through understanding of how decisions are made by 
the machine-constructed models~\cite{jobin2019global}, and what
are the determining factors in these decisions.
One important requirement of machine-constructed models in workflows
with humans-in-the-loop is that the models are human-understandable. Domain-knowledge can be used in two different ways to assist this. First,
it can constrain the kinds of models that are deemed understandable.
Secondly, it can provide concepts that are meaningful for use in
a model. Most of the work in this review has been focused on improving
predictive performance. However, the role of domain-knowledge in constructing
explanations for deep network models is also being explored
(see, for example, \cite{srinivasan2019logical}). However, that
work only generates {\em post hoc\/} explanations that are locally
consistent. Explanatory deep network models that identify true causal connections based on concepts provided as domain-knowledge remain
elusive.

Domain-knowledge can also
be used to correct biases~\cite{mehrabi2021survey} built into
a deep network either declaratively, through the use of constraints,
or through the use of loss functions that include ``ethical penalty''
terms. Demonstrations of the use of domain-knowledge driven,
ethics-sensitive machine learning have been available in
the literature for some time~\cite{Anderson2005MedEthExTA}.
Can these carry over
to the construction of deep network models? This remains
to be investigated.

The issues raised above all go beyond just the ``how''
questions related to the incorporation
of domain-knowledge into deep networks.
They provide pointers to why the use of domain-knowledge
may extend beyond its utility for prediction.

\subsection*{Contributions of Authors}
This paper was conceived and written by TD and AS. SC and
AA contributed to some sections of an earlier version of the paper
(available at: \url{https://arxiv.org/abs/2103.00180}).

\subsection*{Acknowledgements}
AS is the Class of 1981 Chair Professor at BITS Pilani; a
Visiting Professorial Fellow at UNSW Sydney; and a TCS Affiliate Professor.

\bibliographystyle{naturemag}
\bibliography{main}

\end{document}